\documentclass[preprint,12pt]{elsarticle}

\usepackage{amssymb}
\usepackage{amsmath}

\usepackage{color}
\usepackage{tikz}
\usepackage{pgfplots}
\pgfplotsset{compat=1.18}
\usepackage{algorithm}
\usepackage{algorithmic}

\usepackage{url}
\usepackage{graphicx}
\usepackage{caption}
\usepackage{subcaption}
\usepackage{multirow}
\usepackage{stmaryrd}
\graphicspath{{media/}}
\newcommand{\argmax}{\mathop{\rm arg~max}\limits}

\makeatletter
\newcommand{\rev@one}[1]{\textcolor{blue}{#1}}
\long\def\rev@two2#1{\textcolor{red}{#1}}
\DeclareRobustCommand{\rev}{\@ifnextchar2{\rev@two}{\rev@one}}
\makeatother

\journal{BioSystems}

\begin{document}

\begin{frontmatter}



\title{
Satisficing as competitive Bayesian weighting:\\
a minimal model of goal-directed choice under an unattained target
}


\author[address1]{Moto Shinriki}
\author[address1]{Yu Kono}
\author[address1]{Tatsuji Takahashi\corref{cor1}}
\cortext[cor1]{Corresponding author}

\affiliation[address1]{organization={Graduate School of Advanced Science and
Engineering, Tokyo Denki University},
            city={Hatoyama-machi},
            postcode={350-0394}, 
            state={Saitama},
            country={Japan}}
\ead{tatsuji.takahashi@gmail.com}

\begin{abstract}
Animals and humans often do not maximize; they act relative to goals (aspiration levels) that are frequently not yet met.
How can a goal that is not (yet) realized organize choice? Deacon (2012) calls such absence-referenced organization \emph{ententional}, and Garc\'{i}a-Valdecasas and Deacon (2024) argue that ends are best understood as constraints rather than as particular states.
We examine this idea in a minimal formal setting: Risk-sensitive Satisficing (RS), a bandit learning rule that values each action by its share of past trials multiplied by its surplus or shortfall relative to an aspiration level.
We show that RS can be rewritten as a Bayesian-like estimator, Attention Competitive Bayes (ACB), in which the aspiration level is the prior mean of every action and the finite budget of trials sets how strongly evidence can override it.
Two consequences follow.
First, actions that are no longer sampled are partially overridden by the goal rather than by past observations: their estimates relax toward the aspiration level without any data being discarded.
This ``pseudo-forgetting'' distinguishes ACB from models with an explicit forgetting rate.
Second, when the aspiration is unattainable, exploration does not settle; in Bernoulli bandit simulations, adjusting the aspiration level from evidence reduced regret (expected loss).
ACB offers an analyzable model in which an unattained goal and a limited attentional resource jointly constrain behavior, and it yields testable predictions about how organisms revalue options they have abandoned.
\end{abstract}


\end{frontmatter}



\section{Introduction}
\label{sec:introduction}
Living systems rarely pursue a single best outcome.
A foraging animal needs enough energy to survive the night, not the maximum obtainable.
A person searching for an apartment stops searching when one is good enough.
Simon called this \emph{satisficing}: behavior organized with respect to an aspiration level rather than an optimum~\cite{simon_behavioral_1955,simon1956rational}.
Satisficing is not merely a cheaper approximation to optimization.
In studies of risk-sensitive foraging, animals whose expected intake falls short of their requirement tend to prefer variable food sources, whereas those above it tend to prefer reliable ones~\cite{caraco1980empirical,stephens1981logic,bateson2002recent}.
The risk attitude of an organism can thus switch according to whether a goal is currently met.

This raises a basic question about biological organization: 
\emph{how can a goal that is not yet attained shape what an organism does next?}
Under maximization, the target is a single best option, and the fact that it is not yet attained plays no role in the valuation and choice itself.
Under satisficing, the target is a range of acceptable outcomes, rather than a specific value (point), and whether that range has been reached changes behavior.

Deacon called phenomena that are organized with respect to something not (yet) present \emph{ententional}~\cite{deacon2012incomplete}.
His examples include functions that have conditions of satisfaction.
Garc\'{i}a-Valdecasas and Deacon further argued that ends are represented by constraints rather than by particular physical states~\cite{garciavaldecasas2024origins}.
This view suggests a way to approach the above question formally, but doing so requires a minimal model that can be analyzed.

In this paper, we use Risk-sensitive Satisficing ($\mathrm{RS}$), which we proposed in previous works, as such a minimal model
\cite{Takahashi2015,kohno2017cognitive,oyo2017optimization,wakabayashi2020balancing,
shinriki2020flexibility,tsuboya2024reinforcement,tsuboya2025deeprs}.
$\mathrm{RS}$ is a simple value bandit learning rule that values each action by the product of its share of past trials and its surplus or shortfall relative to the aspiration level.
Hereafter, we call this aspiration level the satisfaction level $S$.
An $\mathrm{RS}$ agent optimistically explores rarely tried actions when no action meets the satisfaction level, and it conservatively exploits an action that meets the level when one exists.
Kamiya and Takahashi showed that $\mathrm{RS}$ and Softsatisficing, its softmax version, can serve as models of risk-sensitive decision making such as animal foraging~\cite{kamiya2022softsatisficing}.
Tamatsukuri and Takahashi proved that the cumulative expected loss (hereafter, regret; see Section~\ref{sec:problem_setting_and_related_work} for details) of $\mathrm{RS}$ is bounded by a finite value under specific conditions~\cite{tamatsukuri2019guaranteed}.
However, the theoretical structure behind its defining equation, namely, how a satisfaction level that is not yet attained enters the valuation of actions, has not been sufficiently clarified.

In this paper, we address this issue of what it is to have a goal through three questions.
\emph{First, how does the satisfaction level enter the valuation of actions?}
We reinterpret the defining equation of $\mathrm{RS}$ and show that it has a Bayesian-like structure in which the satisfaction level serves as the prior mean of every action.
We formulate this structure as Attention Competitive Bayes (ACB).
\emph{Second, what does this model predict about actions that are no longer chosen?}
In ACB, the estimates of such actions approach the satisfaction level without discarding data.
We call this \emph{pseudo-forgetting} and compare it with existing Bayesian inference models through simulations.
\emph{Third, what happens when the satisfaction level is unattainable, and how can the level be adjusted?}
We show that exploration does not settle when the satisfaction level is too high, and we examine the effect of adjusting the level from observations, comparing $\mathrm{RS}$ with other satisficing methods.

This study is also related to reinforcement learning (RL).
Traditionally, RL has been treated mainly as an optimization problem aimed at maximizing cumulative rewards~\cite{sutton1998reinforcement}.
This framework has a theoretical foundation inspired by biological behavior and cognition~\cite{Murakoshi2004,murakoshi_simulation_2005,Uragami2013}, and it has achieved remarkable results under certain levels of environmental complexity~\cite{silver_mastering_2017,wurman_outracing_2022}.
RL methods that incorporate human cognitive biases have also been proposed~\cite{Uragami2013,Uragami2016b,ochi_overtaking_2015,kamiura_optimism_2017}.
However, real-world applications of RL remain limited~\cite{abel2024three}.
Ortner et al.\ point out that the cause lies in the gap between RL theory, which aims at optimization, and real-world problems~\cite{ortner_reinforcement_nodate}.
Many real-world problems do not necessarily require finding the optimal solution; it is often more important to quickly find a sufficiently good behavior that achieves the goal.
With this motivation, Michel and Ortner proposed a framework of satisficing RL, in which the agent aims to find actions whose rewards exceed a satisfaction level, together with a method called SAT-UCB~\cite{michel_regret_2022}.
We also place $\mathrm{RS}$ in this context of satisficing RL and compare it with SAT-UCB$^{+}$ (a performance improved variant) and with Thompson Sampling, an optimization method with highest performance.
Our main aim, however, is not to establish the superiority of a method but to answer the three questions above.

The rest of this paper is organized as follows.
Section~\ref{sec:problem_setting_and_related_work} describes the task settings and related work, and Section~\ref{sec:rs} describes the definition of $\mathrm{RS}$ and its exploration process.
Section~\ref{sec:acb_estimation_model} reinterprets $\mathrm{RS}$ as ACB (the first question).
Section~\ref{sec:experiment_acb_vs_existing_bayesian_models} compares pseudo-forgetting with existing Bayesian inference models (the second question).
Section~\ref{sec:compare_satisficing_rl_methods} examines the case in which the satisfaction level is unattainable (the third question).
Section~\ref{sec:discussion} discusses the results as a whole, and Section~\ref{sec:conclusion} concludes the paper.

\section{Goal-Referenced Choice: Task Settings and Related Work}
\label{sec:problem_setting_and_related_work}
The task considered in this paper is a satisficing RL task, specifically the
satisficing bandit task.
In this section, we clarify the task setting considered in this paper.
We also review related work to clarify the position of $\mathrm{RS}$ and its Bayesian interpretation, ACB.

\subsection{Satisficing Bandit Tasks}
\label{subsec:satisficing_bandit_task}
In multi-armed bandit tasks, an agent chooses an action from the action
set $\llbracket 1, K \rrbracket := \{1, 2, \cdots, K\}$.
Hereafter, we call an action in a bandit task an \emph{arm}.
Each arm yields a reward according to a reward distribution that is unknown
to the agent.
At time step $t$, the agent chooses an arm $a_t \in \llbracket 1, K \rrbracket$
and receives the reward as the result of that choice.
In standard (optimization) bandit tasks, the agent repeats this procedure
a finite number of times and aims to maximize the cumulative sum of obtained
rewards.

\begin{figure*}[!tbp]
  \centering
  \includegraphics[width=0.75\linewidth]{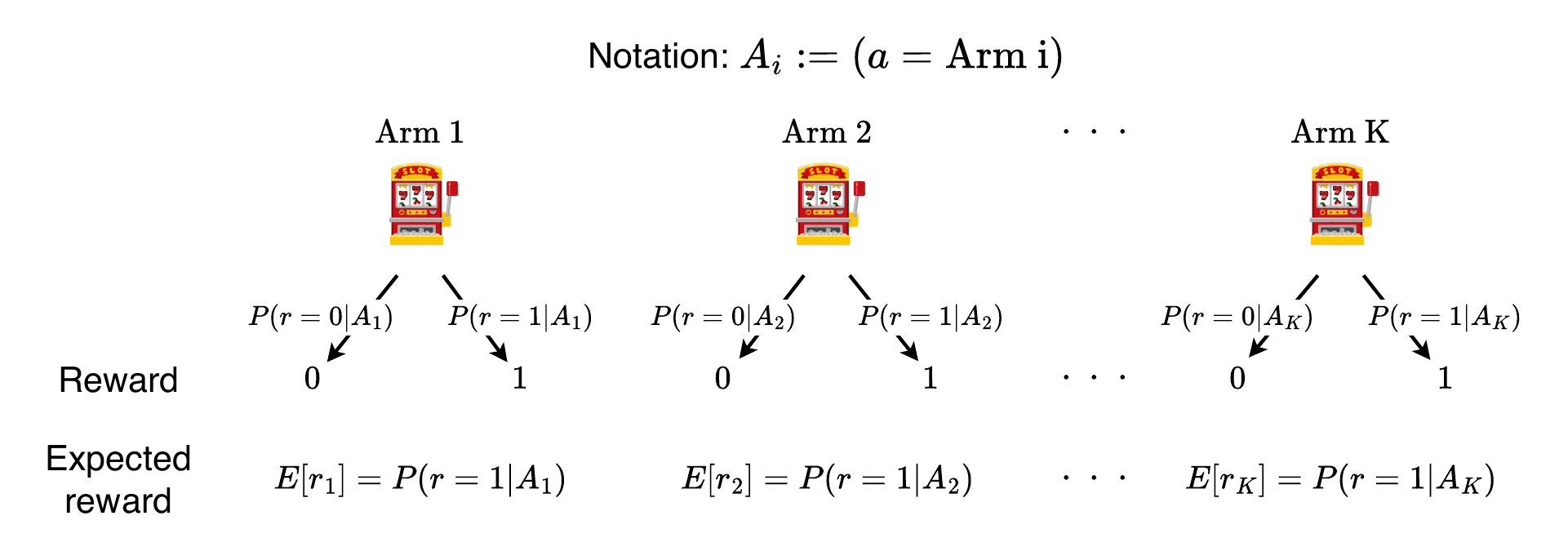}
  \caption{
  The Bernoulli bandit task with $K$ arms.
$E[r_i]$ denotes the expected reward of an arm.
  In this paper, we consider tasks only under this setting.}
  \label{fig:bandit_problems}
\end{figure*}

In contrast, in the satisficing bandit task, the agent's target is not to
maximize cumulative reward, but to choose an arm 
whose reward exceeds a
given satisfaction level $S$, more precisely, whose sample mean reward exceeds $S$
\cite{michel_regret_2022}.
Among satisficing bandit tasks, this paper focuses on Bernoulli bandit
tasks, in which each arm yields a reward $r\in\{0,1\}$ and is
characterized by the probability that $r=1$.
Fig.~\ref{fig:bandit_problems} illustrates an overview of the task setting
considered in this paper.
%
In satisficing bandit task, how the satisfaction level $S$ is set
strongly affects learning behavior and performance.
One such setting is the ``optimal satisfaction level'' $S_{\text{opt}}$.
In this paper, we define the optimal satisfaction level as the midpoint between
the highest true expected reward $E[r_{1\text{st}}]$ and the second-highest
true expected reward $E[r_{2\text{nd}}]$.
Thus, we define $S_{\text{opt}}$ by
\begin{align}
  \label{eq:optimal_satisficing_value}
  S_{\text{opt}} := \frac{E[r_{1\text{st}}] + E[r_{2\text{nd}}]}{2}.
\end{align}
With this satisfaction level, the only arm that satisfies the satisfaction
level is the best arm, and satisficing becomes equivalent to optimization.

In this paper, we use regret as a performance metric for the satisficing bandit
task.
Regret is defined as the cumulative difference between
the true expected reward of the best arm, $E[r_{1\text{st}}]$,
and that of the actually chosen arm at time step $t$, $E[r_{\text{chosen}}]$.
Thus, we define regret by
\begin{align}
  \label{eq:regret_definition}
  \text{regret} := \sum_{t=1}^{T}\left( E[r_{1\text{st}}] - E[r_{\text{chosen}}] \right).
\end{align}
Regret is a metric in optimizing bandit tasks that measures the gap from
the optimal action sequence.
We use regret for the following reasons in the satisficing bandit task.
We adopt Thompson Sampling (TS) \cite{thompson1933likelihood}, a representative
optimizing method \cite{kaufmann2012thompson,agrawal2012analysis}, as the evaluation baseline.
For this reason, we need a metric that enables us to compare satisficing methods
and optimizing methods on the same scale.
It is also reasonable to adopt regret as a common metric for examining how
satisficing-based methods behave when evaluated by an optimizing
metric.
Michel \& Ortner proposed S-regret as a
satisficing-specific metric \cite{michel_regret_2022}.
S-regret is a metric that does not increase when the reward is greater than
the satisfaction level $S$, and increases by the difference between $S$
and the reward when the reward is less than $S$.
For this reason, when the satisfaction level is too high, comparing methods using S-regret reduces to comparing the size of their slopes.
This comparison is not substantially different from comparing where the normal regrets of multiple algorithms converge.
For these reasons, we do not use S-regret as the primary metric, and instead we
use regret, which enables comparison of both satisficing methods and optimizing
methods by the same measure.

\subsection{Related Work}
\label{subsec:related_work}
Studies related to satisficing come from two lines: models of decision making in living systems and methods in RL.
In the former line, Simon formulated satisficing as a theory of decision making under resource constraints~\cite{simon1956rational}.
Studies of risk-sensitive foraging have shown that the risk attitude of animals depends on the relation between their energy budget and their requirement~\cite{caraco1980empirical,stephens1981logic,bateson2002recent}.
Kamiya and Takahashi examined $\mathrm{RS}$ and Softsatisficing as models of such risk-sensitive foraging behavior~\cite{kamiya2022softsatisficing}.
Murakoshi et al.\ used a new RL method to conduct simulations of rat behavior~\cite{murakoshi_simulation_2005}.
Kim et al.\ proposed the TOW model, an amoeba-inspired framework for efficiently solving bandit problems~\cite{Kim2010}, which implicitly performs satisficing.
Uragami et al.\ proposed learning through social information sharing (social satisficing)~\cite{uragami_social_2024}.
We position $\mathrm{RS}$ and ACB in this paper as a minimal model in this line.
Studies in the latter line, on satisficing RL, can be 
broadly classified into the following
two categories: (i) RL methods that aim to achieve satisficing;
and (ii) methods that define satisficing in a different way than the one considered in this paper.

$\mathrm{RS}$ considered in this paper also belongs to category (i) and
is one of the satisficing RL methods proposed by Takahashi et al \cite{Takahashi2015}.
Oyo and Takahashi analyzed the behavior of $\mathrm{RS}$ in bandit problems
\cite{oyo2017optimization}.
Tamatsukuri and Takahashi showed that, when the optimal satisfaction level is given,
the regret of $\mathrm{RS}$ converges in finite time
\cite{tamatsukuri2019guaranteed}.
Furthermore, Tsuboya, Wakabayashi, and Shinriki and others extended $\mathrm{RS}$ to
contextual bandits \cite{tsuboya2024reinforcement}, tabular RL
\cite{wakabayashi2020balancing,shinriki2020flexibility}, and deep RL
\cite{tsuboya2025deeprs}, demonstrating its effectiveness across a wide
range of problem settings.
Kamiya and Takahashi also leveraged properties of $\mathrm{RS}$ and proposed
Softsatisficing as a method for adjusting the inverse temperature (for each arm) of the
Softmax policy \cite{kamiya2022softsatisficing}.
Section~\ref{sec:rs} describes the definition and properties of $\mathrm{RS}$
in detail.
Michel et al.\ proposed SAT-UCB as a satisficing RL method \cite{michel_regret_2022}.
In this paper, we categorize SAT-UCB as a method in category (i).
SAT-UCB is based on the UCB method\cite{auer2002finite}, that is proven to have an optimal property in bandit problems, and widely used in RL-related methods such as AlphaGo \cite{silver2016mastering}.
A SAT-UCB agent switches its action-selection behavior depending on whether the
sample mean reward of arm $i$, $\bar{r}_{i}$, exceeds the satisfaction level
$S$.
If $\bar{r}_{i} \geq S$, the agent selects from arms that satisfy the
satisfaction level.
If $\bar{r}_{i} < S$, the agent performs exploration based on UCB.
In this paper, we show the results of SAT-UCB$^{+}$, a performance-improved variant of SAT-UCB
\cite{michel_regret_2022}, as a comparison method.
Reverdy et al.\ also proposed the UCL algorithm as a Bayesian method for
satisficing RL \cite{reverdy2014modeling,reverdy2016satisficing}.
We categorize UCL in (i).
UCL assumes Gaussian rewards and estimates the expected reward of each arm
by Bayesian inference.
Reverdy et al.\ formulated satisficing in terms of two aspects:
\textit{satisfaction} and \textit{sufficiency}.
In addition to a satisfaction level $M$, which corresponds to $S$ in this paper,
UCL has a sufficiency parameter $\delta \in [0,1]$.
When $\delta>0$, the agent becomes satisfied and stops exploring once
the chosen arm meets the satisfaction level with probability at least $1-\delta$.
In the experiments of Michel et al., UCL performed worse than SAT-UCB$^{+}$
\cite{michel_regret_2022}.
Therefore, we use SAT-UCB$^{+}$ as a benchmark and do not compare our method with UCL.

There are studies whose objectives differ from those of the satisficing RL
considered in this paper, namely, methods categorized as (ii).
Locatelli et al., Mukherjee et al., and Kano et al. studied threshold bandit
tasks \cite{locatelli2016optimal, mukherjee2017thresholding,kano2019good}.
Wu et al.\ studied conservative bandit tasks \cite{wu2016conservative}.
Merlis et al.\ studied lenient regret \cite{merlis2021lenient}.
These studies have some aspects in common with ours, but their problem
settings qualitatively differ.
Russo and Van Roy also investigated satisficing under discounted
rewards and provided a bound on the expected discounted
regret for a satisficing extension of TS \cite{russo2022satisficing}. 
These algorithms are also incomparable and hence out of the scope of the current study. 

\section{A Minimal Model: Risk-sensitive Satisficing ($\mathrm{RS}$)}
\label{sec:rs}
$\mathrm{RS}$ is one of the satisficing RL methods
that incorporates satisficing, a decision-making tendency, into RL.
Unlike the other prior algorithms mentioned above, $\mathrm{RS}$ affects both the agent's action selection and value updating.
This paper tries to clarify the theoretical structure underlying this characteristic, 
in the light of Bayesian belief update.
In this section, we describe the definition of $\mathrm{RS}$ and the exploration
process specific to $\mathrm{RS}$.

\subsection{Definition of $\mathrm{RS}$}
\label{subsec:modify_rs}
Let us consider the multi-armed bandit tasks. 
Let $n_i^{t}$ denote the number of times
that the agent has chosen arm $i \in \llbracket 1, K \rrbracket$ up to
time step $t$.
We define the total number of trials by
\begin{align*}
  N^{t} := \sum_{i=1}^{K} n_i^{t}.
\end{align*}
We define the trial \textit{proportion} of arm $i$ by
\begin{align*}
  \rho_i^{t} := \frac{n_i^{t}}{N^{t}}.
\end{align*}
We define the sample mean reward observed from arm $i$ up to time step $t$ by
\begin{align*}
  \bar r_i^{t} := \frac{1}{n_i^{t}}
  \sum_{s=1}^{t} r_s \,\mathbb{I}\{a_s=i\}
  \qquad (n_i^{t}>0).
\end{align*}
The symbol $\mathbb{I}$ denotes the indicator function.
The $\mathrm{RS}$ agent selects arms based
on the value function $I$ defined as follows:
\begin{align}
  \label{eq:rs}
  I^{t}_{i}
  := \rho_i^{t}\bigl(\bar r_i^{t} - S\bigr).
\end{align}
At each time step $t$, the $\mathrm{RS}$ agent greedily selects an arm:
\begin{align}
\label{eq:rs_action_selection}
  a_t = \argmax_i I^{t}_{i}.
\end{align}
Tamatsukuri \& Takahashi proved that the regret converges to a finite value for
$\mathrm{RS}$ when it uses an optimal satisfaction level $S_{\text{opt}}$ (Eq. \ref{eq:optimal_satisficing_value}) as
the satisfaction level \cite{tamatsukuri2019guaranteed}.

\subsection{Exploration Process Specific to $\mathrm{RS}$}
\label{subsec:exploration_process_rs}
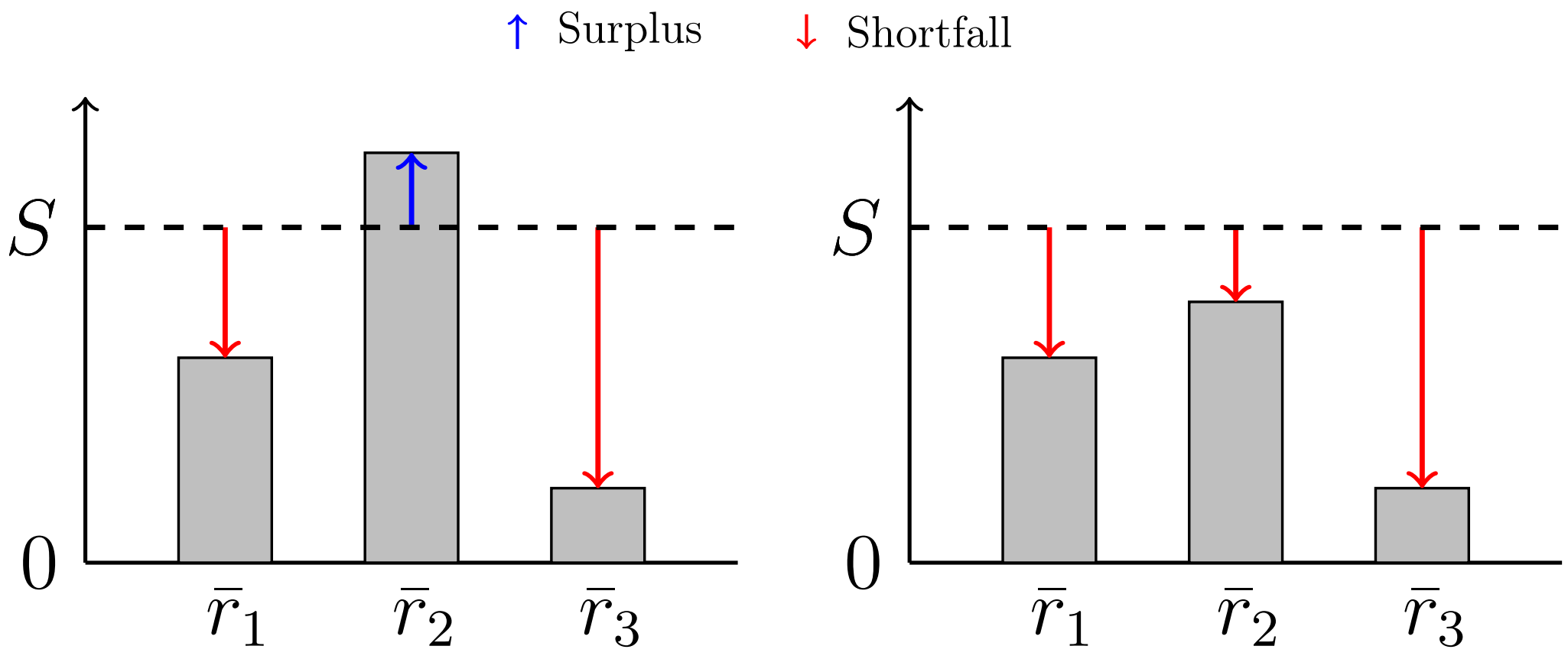
\begin{figure}[tbp]
  \centering
  \begin{tikzpicture}
    \draw[blue,->,thick] (0,0) -- (0,0.3);
    \node[right] at (0.2,0.15) {\small Surplus};
        
    \draw[red,<-,thick] (2.5,0) -- (2.5,0.3);
    \node[right] at (2.7,0.15) {\small Shortfall};
  \end{tikzpicture}

\begin{minipage}[t]{0.48\columnwidth}
    \centering
    \resizebox{\linewidth}{!}{%
    \begin{tikzpicture}
      \draw[semithick,->] (0,-0.5) -- (0,2);
      \draw (0,-0.5) node[left]{0};
      \draw (0,2) node[left]{};
      \draw[semithick] (0,-0.5) -- (3.5,-0.5);
            
      \draw (0.5,-0.5) node[below right]{$\bar{r}_{1}$};
      \fill[lightgray] (0.5,-0.5) rectangle (1,0.6);
      \draw (0.5,-0.5) rectangle (1,0.6);
      \draw[red,->,thick] (0.75,1.3) -- (0.75,0.6);
            
      \draw (1.5,-0.5) node[below right]{$\bar{r}_{2}$};
      \fill[lightgray] (1.5,-0.5) rectangle (2,1.7);
      \draw (1.5,-0.5) rectangle (2,1.7);
      \draw[blue,->,thick] (1.75,1.3) -- (1.75,1.7);
            
      \draw (2.5,-0.5) node[below right]{$\bar{r}_{3}$};
      \fill[lightgray] (2.5,-0.5) rectangle (3,-0.1);
      \draw (2.5,-0.5) rectangle (3,-0.1);
      \draw[red,->,thick] (2.75,1.3) -- (2.75,-0.1);
            
      \draw[thick, dashed] (0,1.3) node[left]{$S$} -- (3.5,1.3);
      \draw[thick, dashed] (0,1.3) node[left]{$S$} -- (3.5,1.3);
    \end{tikzpicture}%
    }
    \subcaption{Over-achieved condition}
    \label{satis}
  \end{minipage}%
  \hfill%
  \begin{minipage}[t]{0.48\columnwidth}
    \centering
    \resizebox{\linewidth}{!}{%
    \begin{tikzpicture}
      \draw[semithick,->] (0,-0.5) -- (0,2);
      \draw (0,-0.5) node[left]{0};
      \draw (0,2) node[left]{};
      \draw[semithick] (0,-0.5) -- (3.5,-0.5);
            
      \draw (0.5,-0.5) node[below right]{$\bar{r}_{1}$};
      \fill[lightgray] (0.5,-0.5) rectangle (1,0.6);
      \draw (0.5,-0.5) rectangle (1,0.6);
      \draw[red,->,thick] (0.75,1.3) -- (0.75,0.6);
            
      \draw (1.5,-0.5) node[below right]{$\bar{r}_{2}$};
      \fill[lightgray] (1.5,-0.5) rectangle (2,0.9);
      \draw (1.5,-0.5) rectangle (2,0.9);
      \draw[red,->,thick] (1.75,1.3) -- (1.75,0.9);
            
      \draw (2.5,-0.5) node[below right]{$\bar{r}_{3}$};
      \fill[lightgray] (2.5,-0.5) rectangle (3,-0.1);
      \draw (2.5,-0.5) rectangle (3,-0.1);
      \draw[red,->,thick] (2.75,1.3) -- (2.75,-0.1);
            
      \draw[thick, dashed] (0,1.3) node[left]{$S$} -- (3.5,1.3);
      \draw[thick, dashed] (0,1.3) node[left]{$S$} -- (3.5,1.3);
    \end{tikzpicture}%
    }
    \subcaption{Under-achieved condition}
    \label{non_satis}
  \end{minipage}
  \caption{Illustration of the two states defined by the relative magnitude of
  the sample mean of rewards, $\bar{r}_{i}$, with respect to the satisfaction level $S$.
  (a) The over-achieved condition occurs if there exists at least one arm $i$
such that $S \leq \bar{r}_{i}$.
  (b) The under-achieved condition occurs if $S > \bar{r}_{i}$ for all arms.}
\end{figure}

$\mathrm{RS}$ agent behaves differently in exploration and exploitation
depending on whether the sample mean of rewards, $\bar{r}^{t}_{i}$, exceeds the satisfaction
level $S$.
In this subsection, we explain the exploration process specific to $\mathrm{RS}$.
To describe the behavior of $\mathrm{RS}$, we distinguish two states, the
over-achieved condition and the under-achieved condition, based on the relative
relationship between the satisfaction level and the sample means
(Figs.~\ref{satis} and \ref{non_satis}, resp.).
%
The over-achieved condition is where there exists an arm $i$ that satisfies $\bar{r}^{t}_{i} \geq S$.
The under-achieved condition is where for all arm $i$, $\bar{r}_i^{t} < S$ holds.
$\mathrm{RS}$ switches its risk attitude depending on these two conditions.
In the over-achieved condition, the agent keeps selecting arms that
satisfy the satisfaction level.
The agent also prefers more \emph{reliable} arms among those that satisfy the
satisfaction level.
That is, arms with larger $\rho_i^{t}$ tend to be preferred.
From Eq.~\eqref{eq:rs}, for arms satisfying $\bar r_i^{t}>S$, a larger
$\rho_i^{t}$ yields a larger $I^{t}_{i}$, which promotes exploitative action
selection by the agent.
In contrast, in the under-achieved condition, $\bar r_i^{t}<S$ holds for all
arms.
In this case, when the trial proportion $\rho_i^{t}$ is smaller, the negative
difference $\bar r_i^{t}-S$ has a smaller effect, and $I^{t}_{i}$ becomes larger.
As a result, the agent tends to evaluate the potential of arms with fewer
trials.
Therefore, in the under-achieved condition, the agent tends to select arms with
fewer trials more, and exploration is promoted.
Such optimistic exploration of $\mathrm{RS}$ in the under-achieved condition is
similar to the principle for the UCB methods, namely, ``optimism in the face of uncertainty'' \cite{munos2014bandits}, though the optimism in additive in UCB while multiplicative in under-achieved RS. 

This state-dependent switching of risk attitudes enables efficient exploration and 
exploitation.
Michel \& Ortner defined the concept of the ``realizable case,'' which is similar to,
but distinct from, the over-achieved condition, in their paper
\cite{michel_regret_2022}.
The realizable case refers to a situation in which for some $i,\; E[r_i] \geq S$
holds, i.e., a situation in which the task itself is satisfiable.
In contrast, being in the over-achieved condition does not necessarily imply that
the task is realizable.
Conversely, even if the task is in the realizable case, it does not necessarily
imply that the agent is in the over-achieved condition, unless in a limit case.

This switching of risk attitude in $\mathrm{RS}$ is consistent with the findings on risk-sensitive foraging described in Section~\ref{sec:introduction}~\cite{bateson2002recent,kamiya2022softsatisficing}.

\section{The Goal as a Kind of Prior: $\mathrm{RS}$ as Attention Competitive Bayes (ACB)}
\label{sec:acb_estimation_model}
In this section, we analyze the internal structure of $\mathrm{RS}$ and clarify
the competitive mechanism of attention resources inherent in $\mathrm{RS}$.
Based on this analysis, we propose Attention Competitive Bayes (ACB).
ACB is a reinterpretation of $\mathrm{RS}$ from a Bayesian perspective.
The purpose of this section is to deepen our understanding of the exploration
behavior and satisficing process of $\mathrm{RS}$ by reinterpreting it as ACB.

\subsection{Attention Mechanism of $\mathrm{RS}$}
\label{subsec:attention_mechanism_rs}
As in Eq.~\eqref{eq:rs}, $\mathrm{RS}$ uses the trial proportion
$\rho^{t}_{i}$ to calculate the value function $I$.
The difference between the sample mean and the satisfaction level,
$\delta^{t}_{i} = \bar{r}^{t}_{i} - S$, is weighted by the trial proportion.
This structure can also be viewed as arms compete for the
finite resource of trials.
That is, the more trials are allocated to an arm, the more importance is
assigned to $\delta^{t}_{i}$; conversely, the fewer trials are allocated to an
arm, the less importance is assigned to $\delta^{t}_{i}$.
In this paper, we call this resource-allocation process, in which the trial
proportion adjusts the weight of $\delta^{t}_{i}$, 
the \emph{Attention competitive mechanism} inherent in $\mathrm{RS}$.
Here, attention does not refer to selective attention in psychology or to attention mechanisms in machine learning; it refers only to the allocation of a finite number of trials among arms.
In the next subsection, we explicitly extract this mechanism and
formalize it as a more general framework.

\subsection{Bayesian-like Structure of $\mathrm{RS}$}
\label{subsec:bayesian_like_structure_rs}
In this subsection, we clarify the mathematical structure of ACB by
reinterpreting the value function of $\mathrm{RS}$ as an estimator of the
expected reward.
Eq.~\eqref{eq:rs_action_selection} shows that action selection depends only on the relative ordering
of the value functions. Therefore, adding (or subtracting) the same constant $S$ to the value
function does not change the action selection result.
Thus, we reinterpret the value function of $\mathrm{RS}$ plus
$S$ as an estimator of the expected reward, denoted by $q_i$, and define
\begin{align}
  q_i
  =
  S + I_i^t
  = 
  S + \rho_i^{t}(\bar r_i^{t}-S).
  \label{eq:attention_bayes}
\end{align}
Eq.~\eqref{eq:attention_bayes} can be transformed into the following
weighted-average form:
\begin{align}
  q_i
  =
  (1-\rho_i^{t})S + \rho_i^{t}\bar r_i^{t}.
  \label{eq:attention_bayes_sum}
\end{align}
Since $\rho_i^{t}=n_i^{t}/N^{t}$,
\begin{align}
  q_i
  =
  S
  +
  \frac{n_i^{t}}{(N^{t}-n_i^{t})+n_i^{t}}
  (\bar r_i^{t}-S)
  \label{eq:attention_bayes_diff}
\end{align}
and it is expressed as above.

For comparison, we consider the standard Beta-Bernoulli model for a Bernoulli
distribution.
Let $\alpha_0$ and $\beta_0$ be the two parameters
of the initial prior distribution.
We define the expectation $\mu_0$ and the confidence $m_0$ of the initial prior
distribution as
\begin{align}
  \mu_0 &:= \frac{\alpha_0}{\alpha_0+\beta_0}, \\
  m_0 &:= \alpha_0+\beta_0.
\end{align}
Then, the expectation of the posterior distribution given observed data is written as
\begin{align}
  q_i
  &=
  \frac{\alpha_0+n_i^{t}\bar r_i^{t}}{\alpha_0+\beta_0+n_i^{t}} \\
  &=
  \mu_0
  +
  \frac{n_i^{t}}{m_0+n_i^{t}}
  (\bar r_i^{t}-\mu_0).
  \label{eq:beta_bernoulli_bayes_estimation_diff}
\end{align}
(Note that $\alpha_0/(m_0+n_i^t)=\mu_0\cdot m_0/(m_0+n_i^t)=\mu_0(m_0+n_i^t-n_i^t)/(m_0+n_i^t)=\mu_0(1-n_i^t/(m_0+n_i^t)$.)
By comparing Eq.~\eqref{eq:attention_bayes_diff} with
Eq.~\eqref{eq:beta_bernoulli_bayes_estimation_diff},
we find that they have the same mathematical structure
under the following correspondence:
\begin{align}
  \mu_0 &= S, \\
  m_0 &= N^{t}-n_i^{t}.
\end{align}
Similar structural correspondences can also be established for the Gamma-Poisson
model and the Gaussian-Gaussian model.
Table~\ref{tab:bayes_parameters_matrix} summarizes the correspondence between
the initial prior expectation $\mu_0$ and the initial confidence $m_0$ for each
model.
\begin{table*}[tbp]
  \centering
  \caption{Expected Value and Confidence of the Initial Prior for Each Model.}
  \label{tab:bayes_parameters_matrix}
  \begin{tabular}{c|cccc} \hline
       & Beta-Bernoulli & Gamma-Poisson & Gaussian-Gaussian & ACB \\ \hline
      Initial Prior Expectation $\mu_{0}$ & $\alpha_{0} / (\alpha_{0}+\beta_{0})$ & $k / \lambda$ & $\mu_{0}$ & $S$ \\
      Initial Confidence $m_{0}$ & $\alpha_{0}+\beta_{0}$ & $\lambda$ & $\sigma^{2}/\sigma^{2}_{0}$ & $N^{t}-n_i^{t}$ \\ \hline
  \end{tabular}
\end{table*}
%

Our analysis suggests that, although ACB is not a strict Bayesian inference model,
it has a Bayesian-like inference structure that integrates observed data and the
prior satisfaction level.

\subsection{Comparison of ACB ($\mathrm{RS}$) and Existing Bayesian Models}
\label{subsec:property_comparison_acb_existing_bayesian_models}
Eq.~\eqref{eq:attention_bayes_sum} shows that the estimate of an unselected arm (with smaller $\rho$)
approaches the expectation $S$ of the initial prior distribution.
Since this behavior appears as if the observed data for unselected arms are
being weakened, but actually not lost, we call it \emph{pseudo-forgetting} in this paper.
A key feature of pseudo-forgetting in ACB is that it is realized by
the relative allocation of attention,
without explicit data deletion or introducing a forgetting rate.
This property is similar to the \emph{sharpening} and \emph{blurring}
properties of LSVR \cite{kohno2017cognitive},
a precursor model to $\mathrm{RS}$
\cite{Takahashi2015}.
Sharpening refers to the property that estimates for frequently tried arms
become ``sharper''.
Blurring refers to the property that estimates for rarely tried arms remain
around the initial values (the individual differences in the values are ``blurred'').
Thus, the pseudo-forgetting behavior of ACB shares features with the
sharpening and blurring properties of LSVR.
Reinterpreting $\mathrm{RS}$ as ACB also makes it possible to identify its
commonalities with LSVR.

Examples of methods with a forgetting property include
Forgetful-Bayes (Dynamic Belief Model) proposed by Zhang et al.\
\cite{zhang_forgetful_2013}
and FQL (Forgetful Q-learning) proposed by Toyama et al.\
\cite{toyama2019reinforcement}.
These methods explicitly attenuate the influence of past data by introducing
a forgetting rate.
In contrast, in ACB, the accumulated data are retained, and
only their relative influence changes.
We call a general Bayesian inference model Standard-Bayes and summarize the
properties of each method in Table~\ref{tab:model_comparison}.
\begin{table*}[tbp]
  \centering
  \caption{Comparison of Properties among Standard-Bayes, Forgetful-Bayes, ACB, and FQL (forgetful Q-learning).}
  \label{tab:model_comparison}
  \begin{tabular}{l|cccc}\hline
    Properties & Standard-Bayes & Forgetful-Bayes & ACB & FQL \\\hline
    Memory Persistence & Yes & No & Yes & No \\
    Belief Relaxation & No & Yes & Yes & Yes \\
    Initial Belief Variability & No & No & Yes & No \\\hline
  \end{tabular}
\end{table*}
%
In Table~\ref{tab:model_comparison}, \textit{Memory persistence} means
that the observed data are not lost.
\textit{Belief relaxation} is that the estimate of an unselected arm
approaches the initial prior distribution.
\textit{Initial belief variability} refers to that the effective prior strength, expressed as a pseudo-count, changes over time.
In ACB, the fixed satisfaction level $S$ corresponds to the prior mean, whereas $m_{0,i}^{t}:=N^{t}-n_i^{t}$ is the effective prior strength for arm $i$.
This property is distinct from variation in either the prior mean or the normalized mixture weight $1-\rho_i^{t}$.
Although these properties can be theoretically inferred from
the defining equations,
we will verify them through simulations
in Sec.~\ref{sec:experiment_acb_vs_existing_bayesian_models}.

\section{Experiment 1: Comparison of ACB and Existing Bayesian Methods}
\label{sec:experiment_acb_vs_existing_bayesian_models}
In this section, we address the second question: how the relatively unchosen arms are treated. 
We conduct simulation experiments on ACB, Standard-Bayes,
Forgetful-Bayes, and Forgetful Q-learning (FQL).
The purpose of these experiments is not to compare their performances, but to visualize
the time evolution of the internal variables maintained by each method and
compare their properties.
Specifically, we aim to empirically confirm the three properties shown in
Table~\ref{tab:model_comparison}: Memory persistence, belief relaxation, and
initial belief variability.
Based on these results, we also discuss how pseudo-forgetting arises in ACB, which provides a Bayesian-like interpretation of $\mathrm{RS}$.

Here, we summarize the comparison methods.
Forgetful-Bayes assumes that environmental statistics change over time and performs
Bayesian inference accordingly.
In this model, the posterior distribution of the reward probability
$\theta^{t}_{i}$ for each arm $i$ is updated as a weighted mixture
of the posterior distribution at time step $t-1$ and
the initial prior distribution, as follows:
\begin{align}
  \label{eq:dbm_prior}
  \text{Pr}(\theta^{t}_{i}
  = \theta \mid S^{t-1}_{i}, F^{t-1}_{i})
  = \gamma q^{t-1}_{i}(\theta) + (1-\gamma)q_{0}(\theta), 
\end{align}
where the weight $\gamma \in [0,1]$.
In Eq.~\eqref{eq:dbm_prior}, $S$ and $F$ denote the numbers of successes and
failures, respectively, and $q_{0}$ denotes the initial prior distribution, which
follows $\text{Beta}(\alpha, \beta)$.
This update rule enables Forgetful-Bayes to discount past data and adapt to a
changing environment.
See \cite{zhang_forgetful_2013} for details.

FQL is a Q-learning algorithm with forgetting, proposed by Toyama et al
\cite{toyama2019reinforcement}.
FQL is a method that extends standard RL algorithms to reproduce memory decay
over time.
For an unselected action $\tilde{a}$, FQL updates its value as follows:
\begin{align}
    \label{eq:forgetting_update}
    Q(\tilde{a}) \leftarrow
    Q(\tilde{a}) + \alpha_{F} (\mu - Q(\tilde{a})).
\end{align}
Here, $\alpha_{F}$ is the forgetting-rate parameter ($0 \le \alpha_{F} \le 1$),
and $\mu$ is the default-value parameter ($0 \le \mu \le 1$) to which
forgetting converges.
See \cite{toyama2019reinforcement} for details.
In this section, we reinterpret the internal state of FQL in a Bayesian-like
manner to enable comparison as a Bayesian inference model.
However, this reinterpretation is a formal framework intended only for comparing 
internal behavior, and it does not claim that FQL performs strict Bayesian
inference.
For details of this reinterpretation,
see \ref{appsec:FQL_bayesian_like_interpretation}.

\subsection{Experiment Settings}
\label{subsec:experiment_1_settings}
We conducted this experiment under the Bernoulli bandit problem setting shown
in Fig.~\ref{fig:bandit_problems}.
We set the number of arms to $K=2$ and call them arm 1 and arm 2, respectively.
We set the reward probability of arm 1 to 0.2 and that of arm 2 to 0.8.
We ran 10 simulations and plotted the mean result.

For ACB, we set the satisfaction level $S$ to 0.6.
For Standard-Bayes and Forgetful-Bayes, we set $\alpha_0=2.4$ and $\beta_0=1.6$ so
that the mean of the initial prior distribution equals 0.6.
For FQL, we set $\mu=0.6$ and set $\alpha_0=\mu/\alpha_F$ and
$\beta_0=(1-\mu)/\alpha_F$.
In ACB and FQL, we set the initial sample mean of all arms to 0.6.
For the initial value of the action selection counts $n_i^{0}$, we initialized
it to $10^{-4}$ as a very small positive value to avoid making the definition
of $\rho_i^{0}$ undefined.
We set the learning rate of FQL to $\alpha=0.1$ and the forgetting rate to
$\alpha_F=0.8$.
Although FQL does not originally require action selection counts, we introduce
them as internal variables in this experiment for Bayesian-like comparison.
The update of the action selection counts $n_i^{t}$ in FQL follows the next equation:
\begin{align}
\label{eq:fql_action_counts_update}
n_i^{t+1} \leftarrow
\begin{cases}
n_i^{t} + \alpha & (i = a_{t}), \\
(1-\alpha_F)n_i^{t} & (i \neq a_{t}).
\end{cases}
\end{align}
For the selected arm, we increment the count weighted by the learning rate $\alpha$.
For unselected arms, we decay their count by a factor of $(1-\alpha_F)$ to match the decay of $\Delta_{t}$ in Eq.~\eqref{eq:delta_t_plus_one} (\ref{appsec:FQL_bayesian_like_interpretation}).

In this experiment, to ensure a fair comparison across all methods, we controlled
the action selection so that arm 1 and arm 2 were chosen alternately for the first 100
steps.
From step 101 onward, each method selected arms using a greedy policy (action selection by argmax operator) based
on its own internal estimates.
In FQL, the agent selects arms based on its forgetful $Q$-values.

\subsection{Results}
\label{subsec:experiment_1_results}
\begin{figure*}[t]
    \begin{minipage}[t]{\linewidth}
        \centering
        \includegraphics[scale=0.8]{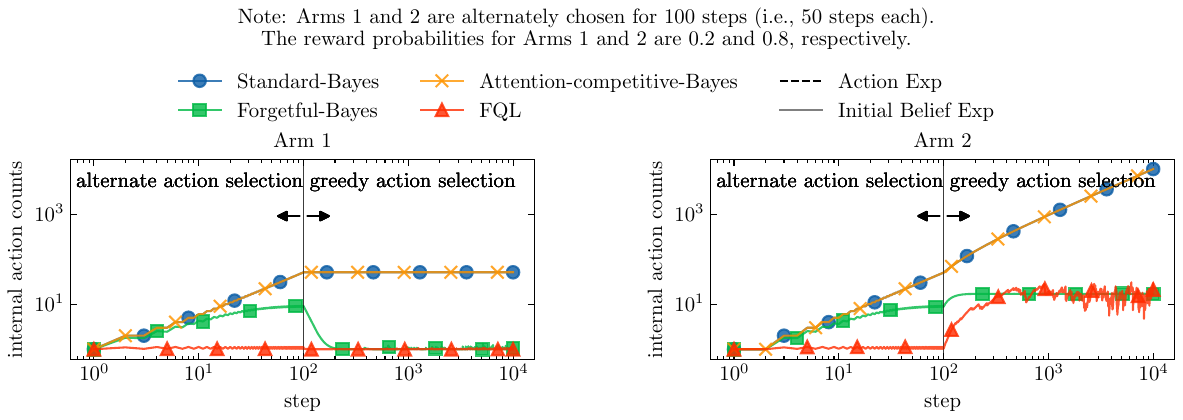}
        \subcaption{Memory persistence}
        \label{fig:memory_persistence}
    \end{minipage}
    \begin{minipage}[t]{\linewidth}
        \centering
        \includegraphics[scale=0.8]{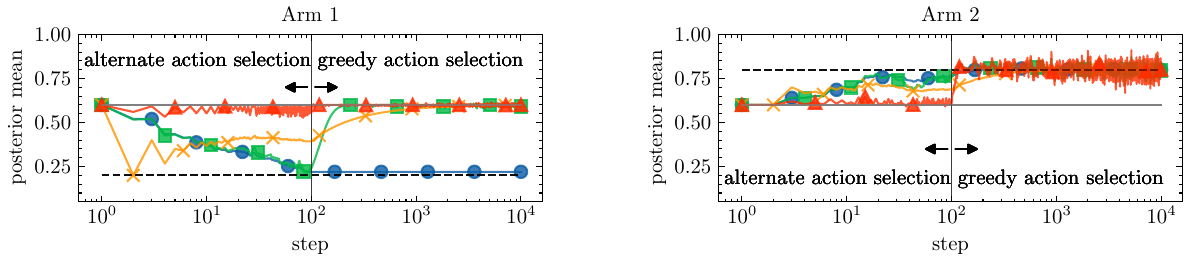}
        \subcaption{Belief relaxation}
        \label{fig:belief_relaxation}
    \end{minipage}
    \begin{minipage}[t]{\linewidth}
        \centering
        \includegraphics[scale=0.8]{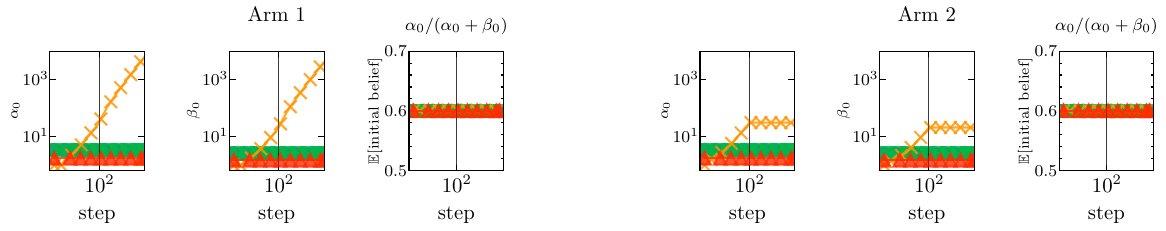}
        \subcaption{Initial belief variability}
        \label{fig:initial_belief_variability}
    \end{minipage}
    \caption{Comparison of internal dynamics among Bayesian-like methods.}
    \label{fig:bayes_compared_results}
\end{figure*}

The results are shown in Fig.~\ref{fig:bayes_compared_results}.
In short, in ACB, the estimate of arm~1, which is abandoned after step 100, returns to the satisfaction level 0.6 without its internal trial count decreasing.
Below, we describe the results for each property: \textit{Memory persistence, belief
relaxation}, and \textit{initial belief variability}.
%

\textbf{Memory persistence}:
Fig.~\ref{fig:memory_persistence} shows memory
persistence, that is, whether the number of trials (or the corresponding
internal variables) that each method maintains internally decays or not.
For Standard-Bayes and Forgetful-Bayes, we plotted $\alpha_t+\beta_t-2$, and
for ACB and FQL, we plotted $n_i^{t}$ as the internal variables on a log--log
scale.
Note that this figure does not show the actual action selection counts but
visualizes the count variables maintained internally by each method.

From step 101 onward, although the selection frequency of arm 1 decreases in all
methods, the way the internal variables change differs substantially across
methods.
In Standard-Bayes and ACB, the internal action counts do not decrease even after
arm 1 is no longer selected.
In contrast, in Forgetful-Bayes, the internal action counts decrease
exponentially as arm 1 stops being selected.
In FQL, due to the effect of the forgetting rate, the action counts fluctuate repeatedly from the early stage of learning.
After the switch to greedy action selection, they hardly increase.
These results confirm that Standard-Bayes and ACB have memory persistence.
They also confirm that Forgetful-Bayes and FQL do not have this property.

\textbf{Belief relaxation}:
Fig.~\ref{fig:belief_relaxation} shows belief
relaxation, that is, the behavior that the estimate of an unselected arm
approaches the expectation of the initial prior distribution.
The solid gray line indicates the mean of the initial prior distribution.
The dashed gray line indicates the true expected reward of each arm.

For arm 2, after the switch to greedy action selection, the estimates
concentrate around the true expected reward in all methods, indicating that the
estimation is appropriate.
For arm 1, a clear difference is observed across these methods.
In Standard-Bayes, even after arm 1 is no longer selected, the posterior
distribution does not change and the estimate remains fixed.
In contrast, in ACB, Forgetful-Bayes, and FQL, the estimates gradually converge
to 0.6, which is the mean of the initial prior distribution.
This behavior indicates that, through true forgetting or pseudo-forgetting
mechanisms, the belief for an arm that receives no new observations is
relaxed toward the initial belief.
From this fact, we confirm that ACB, Forgetful-Bayes, and FQL have the belief
relaxation property.

\textbf{Initial belief variability}:
Fig.~\ref{fig:initial_belief_variability} shows the time evolution of the prior mean and effective prior strength.
We visualize $\alpha_0$, $\beta_0$, and the prior mean $\alpha_0/(\alpha_0+\beta_0)$ for each method.
The prior mean remains constant in all methods, whereas $\alpha_0$ and $\beta_0$, and thus the effective prior strength $\alpha_0+\beta_0$, vary in ACB.
Under the ACB correspondence, $\alpha_{0,i}^{t}=S(N^{t}-n_i^{t})$ and $\beta_{0,i}^{t}=(1-S)(N^{t}-n_i^{t})$.
At each step when arm $i$ is not selected, $N^{t}-n_i^{t}$ increases by one. Thus, both parameters increase linearly while the arm remains unselected.
When arm $i$ is selected, both $N^{t}$ and $n_i^{t}$ increase by one, leaving its effective prior strength unchanged.
This explains why the parameters increase for arm 1 after it is no longer selected, but change little for arm 2 while it continues to be selected.
These results demonstrate initial belief variability in ACB, rather than variation in the prior mean.

\subsection{Discussion}
\label{subsec:experiment_1_discussion}
In this subsection, we discuss the mechanism of pseudo-forgetting in ACB.
As shown in the first-row graph in Fig.~\ref{fig:bayes_compared_results}, in ACB,
even after arm 1 is no longer selected from step 101 onward, the internal trial count
associated with arm 1 does not decrease.
This reflects the fact that ACB does not use an explicit forgetting rate.
On the other hand, in the second row of Fig.~\ref{fig:bayes_compared_results},
we see that the estimate of arm 1 eventually converges to 0.6, which is the mean
of the initial prior distribution.
This means that ACB exhibits pseudo-forgetting, where the belief returns to the
initial prior without discarding past observations.
The reason lies in the change in the confidence of the initial belief.
As shown in the third row of Fig.~\ref{fig:bayes_compared_results}, for arm 1,
which is not selected, the parameters $\alpha_0$ and $\beta_0$ of the initial
prior distribution increase.
This increasing trend means that, for arms that receive no new observations,
the relative influence of the initial belief is strengthened.
Eq.~\eqref{eq:attention_bayes_sum} shows that the weight of the initial belief in
ACB is $1-\rho_i^{t}$, which corresponds to the allocation of attention 
in terms of the trial proportion.
Thus, ACB, and consequently $\mathrm{RS}$, can be interpreted as implementing pseudo-forgetting by changing the weights assigned to data rather than forgetting the data themselves.

\section{Experiment 2: Performance Comparison of Satisficing RL Methods}
\label{sec:compare_satisficing_rl_methods}
In this section, we address the third question: what happens when the satisfaction level is unattainable.
When the satisfaction level is too high, all arms remain in the under-achieved condition, and exploration by $\mathrm{RS}$ does not settle.
We therefore compare, in terms of regret, $\mathrm{RS}$ with a fixed satisfaction level and $\mathrm{RS}$ that adjusts the level from observations.
Comparisons with other satisficing RL methods and an optimizing method are shown as a benchmark check.
We also discuss the results in terms of $\mathrm{RS}$-value dynamics and satisfaction-level adjustment.

We compare the behavior of $\mathrm{RS}$ with SAT-UCB$^{+}$ \cite{michel_regret_2022} and Softsatisficing \cite{kamiya2022softsatisficing}.
We also show the results of Thompson Sampling (TS)\cite{thompson1933likelihood}, a standard optimization algorithm,
as a baseline for comparing satisficing algorithms
.
Michel et al. proposed SAT-UCB$^{+}$ as an experimental algorithm extending the
satisficing RL algorithm SAT-UCB \cite{michel_regret_2022}.
The theoretical guarantee for SAT-UCB$^{+}$ is not yet fully established.
The SAT-UCB$^{+}$ agent uses the given satisfaction level $S$
as the threshold and switches its policy depending on whether
the sample mean reward $\bar{r}^{t}_{i}$ exceeds $S$.
If there are arms that satisfy $\bar{r}^{t}_{i} \geq S$, it selects an arm
among them based on the following equation:
\begin{align}
    \beta^{t}_{i} &= \sqrt{\frac{2\log(f^{t})}{n^{t-1}_{i}}}
    \quad \text{where} \quad f^t = 1 + t \log^{2}(t) \notag \\
    \text{UCB}^{t}_{i} :&= \bar{r}^{t}_{i} + \beta^{t}_{i} \notag ,\quad
    \text{LCB}^{t}_{i} := \bar{r}^{t}_{i} - \beta^{t}_{i} \notag \\
  \label{eq:sat_ucb_plus_action_selection}
  a_t &= \argmax_{i} \Biggl\{
    \frac{ \text{UCB}^{t}_{i} - \max \{S, \text{LCB}^{t}_{i}\} }
    {\beta^{t}_{i}}
    \Biggr\}
\end{align}
In Eq.~\eqref{eq:sat_ucb_plus_action_selection}, UCB and LCB are the upper and
lower confidence bounds, respectively, and $\beta_{i}(t)$ is the confidence
width.
If no satisfactory arm exists, it selects arms in the same way as the
standard UCB1.
The SAT-UCB$^{+}$ agent selects each arm at least once before 
choosing arms according to
Eq.~\eqref{eq:sat_ucb_plus_action_selection}.
See \cite{michel_regret_2022} for details.

Softsatisficing, proposed by Kamiya and Takahashi, is a probabilistic transformation
of $\mathrm{RS}$ \cite{kamiya2022softsatisficing}.
While the $\mathrm{RS}$ agent selects arms deterministically, the
Softsatisficing agent selects arms according to a certain Softmax probability
distribution, which allows its exploration behavior to be represented.
Softsatisficing defines the satisfaction degree of each arm as
$\delta_{i} = \bar{r}^{t}_{i} - S$ and the satisficing value as
$E^{\mathrm{RS}}_{i} = - \log (S - \bar{r}^{t}_{i})$.
Using this $E^{\mathrm{RS}}_{i}$, it defines the action selection probability as
\begin{align}
  \label{eq:softsatisficing}
  \text{Softsatisficing}(a_{i};S)
  = \frac{\exp(E^{\mathrm{RS}}_{i})}
  {\sum_{j}\exp(E^{\mathrm{RS}}_{j})}
\end{align}
and then selects arm according to this probability.
See \cite{kamiya2022softsatisficing} for details.

\subsection{Experiment Settings}
\label{subsec:compare_satisficing_rl_methods_settings}
We conducted this experiment under the Bernoulli bandit problem setting shown
in Fig.~\ref{fig:bandit_problems}.
We set the number of arms to $K=8$.
The reward probability of each arm is sampled uniformly from
$[0,1]$.
We ran 1,000 simulations and plotted the mean regret.
As the satisfaction level $S$, we used the following two values:
\begin{itemize}
  \item the optimal satisfaction level $S_{\text{opt}}$ (see Eq. \ref{eq:optimal_satisficing_value}),
  \item the upper bound on the task reward 1.0
\end{itemize}
We conducted the experiment with 1.0 to examine how each
method behaves when the satisfaction level is set to a relatively high value.

For $\mathrm{RS}$, we set the initial sample mean of all
arms to 0.5.
We initialized the initial action selection count $n_i^{0}$ with a very small
positive value $10^{-4}$ to avoid making the definition of the trial proportion
$\rho_i^{0}$ undefined.
In this experiment, we used both the standard $\mathrm{RS}$ and $\mathrm{RS}$
equipped with a mechanism for dynamically adjusting the satisfaction level,
namely, dynamically adjusted $\mathrm{RS}$.
This dynamic satisfaction-level adjustment mechanism exploits properties of
$\mathrm{RS}$. The reasons for introducing this mechanism and its derivation
are described in
\ref{appendix:adjustment_satisfaction_level}.
Algorithm~\ref{alg:satisfaction_level_adjustment} summarizes the adjustment procedure.
With this mechanism, the satisfaction level of dynamically adjusted $\mathrm{RS}$ can change during a simulation.
We denote the satisfaction level of standard $\mathrm{RS}$ by $S$,
and the original satisfaction level of
dynamically adjusted $\mathrm{RS}$ by $S_{0}$.
Algorithm~\ref{alg:satisfaction_level_adjustment} assumes that the empirically best arm $g$ is unique and that its sample mean falls short of the initial satisfaction level ($\bar r_g<S_{0}$).
The input is the initial satisfaction level $S_{0}$, not the result of the previous adjustment.
At each step $t$, the procedure uses the current counts $n_i^{t}$ and sample means $\bar r_i^{t}$. RS then compares arms using the resulting adjusted satisfaction level $S^{\prime}$.
The algorithm omits the superscript $t$.
\begin{algorithm}[tbp]
\caption{Upper-limit-based satisfaction-level adjustment}
\label{alg:satisfaction_level_adjustment}
\begin{algorithmic}[1]
\REQUIRE Initial satisfaction level $S_{0}\in[0,1]$, counts $n_i>0$, and means $\bar r_i\in[0,1]$ for $i=1,\ldots,K$, with $K\geq2$
\ENSURE Adjusted satisfaction level $S^{\prime}$
\STATE $g\leftarrow\argmax_i\bar r_i$ \COMMENT{unique by assumption}
\FOR{each $j\in\{1,\ldots,K\}\setminus\{g\}$}
  \STATE $D_j\leftarrow D_{\mathrm{KL}}(\operatorname{Ber}(\bar r_j)\,\|\,\operatorname{Ber}(\bar r_g))$
  \STATE $\tilde\tau_j\leftarrow\exp(-n_jD_j)$
  \STATE $S_j\leftarrow\bar r_g+\dfrac{\tilde\tau_j}{1-\tilde\tau_j}(\bar r_g-\bar r_j)$
\ENDFOR
\STATE $S_{\mathrm{UL}}\leftarrow\max_{j\ne g}S_j$
\STATE $S^{\prime}\leftarrow\min\{S_{0},S_{\mathrm{UL}}\}$
\RETURN $S^{\prime}$
\end{algorithmic}
\end{algorithm}
$D_{\mathrm{KL}}(\cdot\|\cdot)$ denotes the KL divergence, and $\operatorname{Ber}(p)$ denotes the Bernoulli distribution with mean $p$.
%
For TS, we set the parameters of the initial Beta distribution to
$\alpha_{0}=1.0$ and $\beta_{0}=1.0$.

\subsection{Results}
\label{subsec:compare_satisficing_rl_methods_results}
\begin{figure*}[tbp]
  \centering
  \begin{minipage}[t]{0.48\textwidth}
    \centering
    \includegraphics[width=\linewidth]{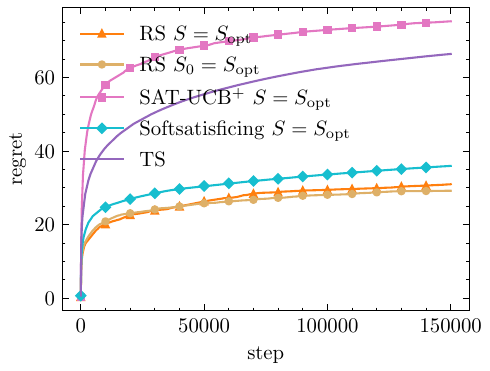}
    \subcaption{Regret with $S=S_{\text{opt}}$}
    \label{fig:opt_regret}
  \end{minipage}%
  \hfill%
  \begin{minipage}[t]{0.48\textwidth}
    \centering
    \includegraphics[width=\linewidth]{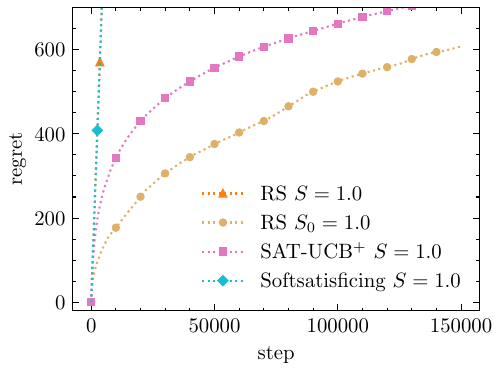}
    \subcaption{Regret with $S=1.0$}
    \label{fig:10_regret}
  \end{minipage}
  \caption{Performance comparison among satisficing RL methods.
  The plot of TS is shown as a baseline.
  In both settings, dynamically adjusted $\mathrm{RS}$ achieves lower regret than SAT-UCB$^{+}$.
The legend $\mathrm{RS}$ $S=x$ denotes standard $\mathrm{RS}$ with the
satisfaction level fixed at $x$, whereas $\mathrm{RS}$ $S_0=x$ denotes
$\mathrm{RS}$ whose satisfaction level is dynamically adjusted from
the initial value $x$.
  (a) Regret over time with $S=S_{\text{opt}}$ or $S_{0}=S_{\text{opt}}$.
  (b) Regret over time with $S=1.0$ or $S_{0}=1.0$.}
\end{figure*}

We show the results in Fig.~\ref{fig:opt_regret} and Fig.~\ref{fig:10_regret}.
%
Fig.~\ref{fig:opt_regret} shows the time evolution of regret when the
satisfaction level is set to $S_{\text{opt}}$ (or $S_{0}=S_{\text{opt}}$).
Under this setting, $\mathrm{RS}$ and Softsatisficing achieve lower regret than
SAT-UCB$^{+}$.
Fig.~\ref{fig:10_regret} shows the results when the satisfaction level is
set to 1.0.
Even in this case, dynamically adjusted $\mathrm{RS}$
achieves lower regret than SAT-UCB$^{+}$.
These results show that, in the settings we tested, satisfaction-level adjustment mitigates the effects of an excessively high initial satisfaction level in $\mathrm{RS}$.
By contrast, standard $\mathrm{RS}$ and Softsatisficing do not
have an adjustment mechanism for the satisfaction level,
and its regret linearly increases over time.
The plots for standard $\mathrm{RS}$ and Softsatisficing overlap almost
completely.

\subsection{Discussion}
\label{subsec:compare_satisficing_rl_methods_discussion}
In the current definition, the RS value is normalized by the trial proportion $\rho_i=n_i/N$ and is written as $I_i=\rho_i(\bar r_i-S)$.
In the original proposal of $\mathrm{RS}$, the value was not normalized by the trial proportion and was given by $n_i(\bar r_i-S)$~\cite{Takahashi2015}.
We now derive the value corresponding to this original definition using the current notation.
Multiplying the current definition by $N$ gives
\begin{align}
    N I_i
    &= N\rho_i(\bar r_i-S) \notag \\
    &= N\frac{n_i}{N}(\bar r_i-S) \notag \\
    &= n_i(\bar r_i-S).
\end{align}
Henceforth, we write this unnormalized value as $N I_i$
to make clear that it is the product of $N$ and $I_i$.
In this discussion, we examine how this value changes under a fixed satisfaction level $S$.
At any given time, $N>0$ is common to all arms, so $N I_i$ and $I_i$ rank the arms identically and yield the same action choice.
Let $I_i^{+}$ denote the normalized RS value after selecting arm $i$ once and receiving reward $r$.
Since the total trial count also increases from $N$ to $N+1$, the increment in the unnormalized value $N I_i$ is
\begin{align}
 \Delta(N I_i):=(N+1)I_i^{+}-N I_i=r-S.
\end{align}
Taking the expectation over the reward, conditional on the state before selection and on selecting arm $i$, gives
\begin{align}
 E[\Delta(N I_i)]=E[r_i]-S.
\end{align}
Here, $\Delta(N I_i)$ denotes the change in the entire product and differs from $N\Delta I_i$.
Thus, the expected value of $N I_i$ increases after selecting arm $i$ if $E[r_i]>S$ and decreases if $E[r_i]<S$.
If the satisfaction level lies between $E[r_{1\text{st}}]$ and
$E[r_{2\text{nd}}]$, namely,
$E[r_{2\text{nd}}] < S < E[r_{1\text{st}}]$, selecting the best arm
increases its $N I_i$ on average.
In contrast, the unnormalized values $N I_i$ of suboptimal arms decrease on average as those
arms continue to be selected.
$\mathrm{RS}$ can autonomously escape from situations
in which it continues to select suboptimal arms.
Tamatsukuri \& Takahashi showed that, under this condition, the regret of
$\mathrm{RS}$ is bounded above by a finite value~\cite{tamatsukuri2019guaranteed}.

The lower regret of standard $\mathrm{RS}$ in Fig.~\ref{fig:opt_regret} is consistent with the theoretical property established above for a fixed satisfaction level.
Dynamically adjusted $\mathrm{RS}$ also achieved low regret in this figure, although this empirical result is separate from the guarantee for a fixed satisfaction level.
In contrast, for standard $\mathrm{RS}$ in
Fig.~\ref{fig:10_regret}, 1.0 is at least as large as $E[r_{1\text{st}}]$.
Hence, $E[r_i] - 1.0 \leq 0$ for all arms.
In this under-achieved setting, even the unnormalized value $N I_i$ of
the best arm does not increase on average.
Therefore, the condition for the finite-regret result described above
is not satisfied.
As shown in Fig.~\ref{fig:10_regret}, the regret of standard $\mathrm{RS}$
rapidly increased in a linear way, overlapping with that of Softsatisficing.

The dynamically adjusted $\mathrm{RS}$ balances exploration and exploitation by lowering an excessively high satisfaction level based on observed data.
The adjusted satisfaction level need not lie between the true mean rewards of the best and second-best arms, and changing the satisfaction level also changes $N I_i$.
Therefore, the expected-increment argument and finite-regret guarantee for a fixed satisfaction level do not directly apply to the dynamic adjustment mechanism.
In Fig.~\ref{fig:10_regret}, dynamic adjustment yielded lower regret than keeping the satisfaction level fixed at 1.0.
These results show empirically that, in the tested setting, satisfaction-level adjustment can curb excessive exploration and reduce regret.

\section{Discussion}
\label{sec:discussion}
In this section, we summarize the answers to the three questions in Introduction and discuss their implications and limitations.

\subsection{The Answers and Two Constraints on Choice}
\label{subsec:two_constraints}
The reinterpretation as ACB shows that action selection in $\mathrm{RS}$ is determined by a combination of two constraints.
One is the satisfaction level $S$, and the other is the finite allocation of trials, $\sum_i\rho_i^{t}=1$.
The fewer trials an arm has received, the more its estimate is described by the satisfaction level rather than by observations (Section~\ref{subsec:bayesian_like_structure_rs}).
The three results of this paper follow from this single structure.
First, in the under-achieved condition, rarely tried arms are valued highly because of the satisfaction level, which promotes exploration (Section~\ref{subsec:exploration_process_rs}).
Second, the estimates of arms that are no longer selected return to the satisfaction level without discarding data (Section~\ref{sec:experiment_acb_vs_existing_bayesian_models}).
Third, when the satisfaction level is unattainable, exploration does not settle, and adjusting the level from observations reduced regret in the tested settings (Section~\ref{sec:compare_satisficing_rl_methods}).
Maximization can be positioned as a special case of this framework.
When the satisfaction level lies between the expected rewards of the best and second-best arms, only one arm satisfies it (Section~\ref{subsec:satisficing_bandit_task}).

\subsection{Relation to Deacon's Framework}
\label{subsec:relation_deacon}
As noted in Section~\ref{sec:introduction}, Deacon called phenomena organized with respect to something not yet present \emph{ententional}, and Garc\'{i}a-Valdecasas and Deacon argued that ends are represented by constraints rather than by particular states~\cite{deacon2012incomplete,garciavaldecasas2024origins}.
The satisfaction level $S$ of $\mathrm{RS}$ specifies a range of acceptable outcomes rather than a single best arm.
Moreover, in the under-achieved condition, $S$ orders the arms even though no arm has reached it.
In this sense, $\mathrm{RS}$ can be regarded as a minimal formal model of action selection referenced to a constraint that is not yet realized.
However, $S$ is given from outside the model.
The model therefore captures the logical form of such organization, not its origin.
In Deacon's framework, the origin of goal-directedness requires a system that maintains the constraints on which its own persistence depends, and a bandit agent does not have this property.

\subsection{Possible Biological Implications}
\label{subsec:biological_implications}
Because we have not compared the model with behavioral data, we state the following as hypotheses.
First, pseudo-forgetting yields a testable prediction about how animals revalue options that they have abandoned.
In ACB, the estimate of an arm that is no longer selected approaches neither zero nor its last observed value, but the current satisfaction level.
In contrast, Forgetful-Bayes approaches the mean of the initial prior, and FQL approaches the default value $\mu$; both values are independent of the satisfaction level~\cite{zhang_forgetful_2013,toyama2019reinforcement}.
In Experiment~1, we set all of these values to 0.6, so this difference did not appear.
A behavioral experiment that manipulates only the aspiration level, for example through deprivation or energy requirements as in studies of risk-sensitive foraging~\cite{bateson2002recent}, might distinguish these models.
Note, however, that relaxation in ACB advances with trials on other arms, not with the passage of time alone.

Second, weighting by the trial proportion $\rho_i=n_i/\sum_k n_k$ has the simplest form of divisive normalization, which is regarded as a canonical neural computation~\cite{carandini2012normalization}.
In value representation, value signals have also been reported to be divisively normalized by the values of the available options~\cite{louie2011reward,louie2012efficient}.
In $\mathrm{RS}$, however, what is normalized is experience (the number of trials), not value, and whether this formal similarity reflects a shared mechanism remains an open empirical question.

Third, the satisfaction-level adjustment in \ref{appendix:adjustment_satisfaction_level} modifies one constraint, the satisfaction level, by another rule based on observations.
It can be seen as a simple example of regulating a constraint, although the rule itself is designed.

\subsection{Limitations and Future Work}
\label{subsec:limitations}
This study has several limitations.
First, the satisfaction level $S$ is given exogenously.
Linking $S$ to the agent's own viability, for example to its energy budget, is one direction toward the self-maintaining case that Deacon considers.
Second, the experiments and the adjustment in the appendix are limited to Bernoulli rewards and stationary environments.
The correspondence of ACB also holds for the Gamma-Poisson and Gaussian-Gaussian models (Table~\ref{tab:bayes_parameters_matrix}), but we have not examined the behavior in those cases.
Third, 
$\mathrm{RS}$ selects arms deterministically, and we have not fitted it to behavioral data.
A stochastic version such as Softsatisficing is more suitable for such comparisons~\cite{kamiya2022softsatisficing}.
Lastly, the satisfaction level adjusment method in Algorithm~\ref{alg:satisfaction_level_adjustment}, described in \ref{appendix:adjustment_satisfaction_level}, is conceptually described but not yet mathematically guaranteed to be optimal. 

\section{Conclusion}
\label{sec:conclusion}
We asked how a goal that is not yet attained can organize choice, and we examined this question in a minimal model, Risk-sensitive Satisficing ($\mathrm{RS}$).
By reinterpreting $\mathrm{RS}$ as Attention Competitive Bayes (ACB), we showed that the satisfaction level enters the valuation of every arm as its prior mean, and that the finite allocation of trials determines how far observations override it.
Two constraints, what counts as good enough and how many trials are available, thus jointly shape behavior.
Three consequences follow from this structure.
Rarely tried arms are explored because they are described by the unattained satisfaction level.
The estimates of abandoned arms return to the satisfaction level without discarding data (pseudo-forgetting).
When the satisfaction level is unattainable, exploration does not settle; in the tested Bernoulli bandit settings, adjusting the level from observations reduced regret.
In the light of Deacon's framework, $\mathrm{RS}$ can be regarded as a minimal formal model of action selection referenced to a constraint that is not yet realized.
Because the satisfaction level is given from outside, however, the model only captures the form of such organization, and not its origin.
Future work should derive the satisfaction level from the agent's own viability, test the prediction of pseudo-forgetting in animal and human choice, and extend the analysis beyond Bernoulli rewards.

\section{Declaration of generative AI and AI-assisted technologies in the manuscript preparation process}

During the preparation of this work, M. Shinriki used Codex (OpenAI) and Claude Code (Anthropic) in order to translate the part of the manuscript he authored from Japanese into English.
After using these tools, the authors thoroughly reviewed and edited the content as needed and take full responsibility for the content of the published article.

\section{Acknowledgment}
This work was partially supported by 
JSPS KAKENHI Grant Numbers 23H03469, 23K28159, and 20H04259, 
and Research Institute for
Science and Technology of Tokyo Denki University
Grant Number Q26D-03 / Japan.

\bibliographystyle{elsarticle-num}  
\bibliography{references}

\appendix

\section{A Bayesian-like Interpretation of FQL as Sequential Updating}
\label{appsec:FQL_bayesian_like_interpretation}
In Sec.~\ref{sec:experiment_acb_vs_existing_bayesian_models}, we compare the
methods from the viewpoint of Bayesian inference models.
Under this comparison, FQL must also be interpreted as a Bayes-like inference model.
In this subsection, we show that although FQL is not strictly a Bayesian
inference model, its update structure closely resembles that of
sequential Bayesian updating.
This interpretation provides a supplementary perspective for understanding FQL
in relation to Bayesian inference models.

In FQL, a forgetting update that decays past value estimates is applied to
unchosen actions $\tilde{a}$.
The update rule is written as:
\begin{align}
  \label{eq:q_learning_update}
  Q_{t+1}(\tilde{a})
  &\leftarrow
  Q_{t}(\tilde{a}) + \alpha_{F} (\mu - Q_{t}(\tilde{a})) \\
  \label{eq:q_learning_update_changed}
  &=
  (1 - \alpha_{F}) Q_{t}(\tilde{a}) + \alpha_{F} \mu.
\end{align}
Here, $\alpha_F \in (0,1)$ is the forgetting rate, and $\mu$ is the default value to
which forgetting converges.
$Q_{t}(\tilde{a})$ is the action value of action $\tilde{a}$ at time step $t$, and
in FQL, action selection is performed based on this $Q$ value.

For a formal comparison with Bayesian inference, we express $Q_{t}$ in terms of
the deviation from the initial value $Q_{0}$.
In the experiment in
Sec.~\ref{sec:experiment_acb_vs_existing_bayesian_models}, the initial value
$Q_{0}$ is set the default value $\mu$, and thus we assume
$Q_{0} = \mu$ in this subsection.
Defining the deviation between $Q_{t}$ and the initial value as
\begin{align}
  \label{eq:init_diff_q0_qt}
  \Delta_{t} := Q_{t} - Q_{0},
\end{align}
we can write $\Delta_{t+1}$ as follows:
\begin{align}
  \Delta_{t+1}
  &= Q_{t+1} - Q_{0} \notag \\
  &= (1 - \alpha_{F})(Q_{t} - Q_{0})
     + \alpha_{F}(\mu - Q_{0}) \notag \\
  \label{eq:delta_t_plus_one}
  &= (1 - \alpha_{F}) \Delta_{t} .
\end{align}
From this recurrence relation, $\Delta_{t}$ is given by
\begin{align}
  \label{eq:delta}
  \Delta_{t}
  =
  (1 - \alpha_{F})^{t-1} \Delta_{1}.
\end{align}
That is, the forgetting update in FQL exhibits exponential regression toward $Q_{0}$.
This exponential decay is structurally similar to the behavior of the corresponding
update equation in sequential Bayesian inference, where the relative influence of
the prior distribution weakens as observations accumulate.

Next, we consider the correspondence between FQL and Bayesian inference in
the Beta-Bernoulli model.
Let $q_{t-1}$ be the mean of the prior distribution and $q_{t}$ be
the mean of the posterior distribution.
When a successful reward ($1$) is observed, the update is given as follows:
\begin{align}
  q_{t}
  &= \frac{\alpha_{t} + 1}{\alpha_{t} + \beta_{t} + 1} \notag \\
  \label{eq:bayes_prior_posterior_mean_diff}
  q_{t} - q_{t-1}
  &=
  \frac{1}{\alpha_{t} + \beta_{t} + 1}
  (1 - q_{t-1}) .
\end{align}
In the context of RL, this expression can be generalized as
\begin{align}
  \label{eq:bayes_prior_posterior_mean_diff_generalization}
  q_{t} - q_{t-1}
  =
  \frac{1}{\alpha_{t} + \beta_{t} + 1}
  (r - q_{t-1}).
\end{align}
Comparing Eq. \eqref{eq:q_learning_update} and
Eq.\eqref{eq:bayes_prior_posterior_mean_diff_generalization}, the forgetting rate
$\alpha_F$ in FQL can be associated with the learning-rate term in Bayesian inference
as follows:
\begin{align}
  \label{eq:q_bayes_analogy}
  \alpha_{F}
  &\longleftrightarrow
  \frac{1}{\alpha_{0} + \beta_{0} + 1} .
\end{align}
Thus, the forgetting update in FQL can be interpreted as assigning unselected actions a pseudo-reward of $\mu$ with learning rate $\alpha_F$.
When $\alpha$ and $\beta$ are sufficiently large, the effect of
the constant term 1 can be neglected, and thus we can approximate
\begin{align}
  \label{eq:q_learning_rate_analogy}
  \alpha_{F}
  \approx
  \frac{1}{\alpha_{0} + \beta_{0}}.
\end{align}
Note that this correspondence is not a strict equivalence but an approximate
analogy in the limit of a large number of trials.

From the above, we can see that the forgetting update in FQL has an update
structure that is formally similar to the weighted updates in sequential
Bayesian inference, in that it exhibits exponential regression toward the
initial value.
In this paper, based on this Bayesian-like interpretation, we position FQL as
one of the comparison methods in the experiments in
Sec.~\ref{sec:experiment_acb_vs_existing_bayesian_models}.

\section{Upper-Limit-Based Dynamic Adjustment of the Satisfaction Level}
\label{appendix:adjustment_satisfaction_level}

This appendix explains the functionality and the mechanism of the dynamic adjustment of the satisfaction level used by dynamically adjusted $\mathrm{RS}$ in Sec.~\ref{subsec:compare_satisficing_rl_methods_settings}.
When the satisfaction level remains unattainable, RS continues exploring toward an equilibrium in which trial proportions compensate for differences in how far sample means fall below the satisfaction level~\cite{kamiya2022softsatisficing}.
Such exploration is reasonable for satisficing, but it can be excessive when one would prefer to exploit higher-reward arms even though the satisfaction level is unattainable.
This appendix presents the method used in Exp. 2 (the results in Fig~\ref{fig:opt_regret} and \ref{fig:10_regret} for RS with $S_0$) for using empirical information to lower the satisfaction level and adjust the balance between exploration and exploitation.
Algorithm~\ref{alg:satisfaction_level_adjustment} in Sec.~\ref{subsec:compare_satisficing_rl_methods_settings} shows the adjustment procedure.
The following subsections explain Algorithm~\ref{alg:satisfaction_level_adjustment} line by line.

\subsection{Setting, Assumptions, and the Empirically Best Arm}
Algorithm~\ref{alg:satisfaction_level_adjustment} takes the initial satisfaction level $S_{0}$, the trial counts $n_i$, and the sample means $\bar r_i$ as inputs, and it outputs the adjusted satisfaction level $S^{\prime}$.
We consider Bernoulli rewards and an original satisfaction level $S_{0}\in[0,1]$.
We describe the adjustment at a fixed time step $t$ and omit the superscript $t$ for simplicity.
We use the RS value $I_i=\rho_i(\bar r_i-S)$ and trial proportion $\rho_i=n_i/\sum_k n_k$ defined in the main text.
In the numerical implementation, trial counts are initialized to $10^{-4}$ (to avoid indeterminate $\rho_i$ values from division by zero) and sample means to $0.5$.
Line~1 of Algorithm~\ref{alg:satisfaction_level_adjustment} selects the arm with the largest sample mean.
Let $g=\argmax_i\bar r_i$ denote the empirically best arm, let $J=\llbracket1,K\rrbracket\setminus\{g\}$ be the set of all other arms, and define $\tau_j=n_j/n_g=\rho_j/\rho_g$.
For the following derivation, we assume that the empirically best arm is unique and that $\bar r_g<S_{0}$.
Let $S^{\prime}$ denote the adjusted satisfaction level and $I_i^{\prime}:=\rho_i(\bar r_i-S^{\prime})$ the changed RS value.
We hold the observed data and trial proportions fixed to examine how changing the satisfaction level affects the current arm comparison.
Setting $S^{\prime}=\bar r_g$ gives $g$ an RS value of zero and every other arm a negative value. Thus, $g$ is the sole maximizer in the current comparison.
Since $g$ is not always the optimal arm, we restrict the adjustment to $S^\prime \in (\bar r_g, S_{0}]$.

\subsection{Quantifying Evidence Against Each Arm Having a Higher True Mean}
Line~2 iterates over all arms $j\in J$ other than $g$. Line~3 computes the KL term $D_j$ for each pair $(j,g)$.
We need a criterion for deciding whether to exploit the current empirically best arm $g$ or further explore another arm $j$.
This decision should account for both the difference in sample means and the amount of evidence supporting it.
We therefore draw on the IMED algorithm, a bandit policy that computes an index for each arm from the observed data and selects an arm with the smallest index~\cite{honda2015non}.
The index is a score for deciding which arm should be chosen, rather than a reward prediction.
IMED constructs this index from the KL divergence between Bernoulli distributions.
We abbreviate the KL divergence between Bernoulli distributions with means $p$ and $q$ as $d(p,q):=D_{\mathrm{KL}}(\operatorname{Ber}(p)\,\|\,\operatorname{Ber}(q))$.
It is given by
\begin{align}
 d(p,q)=p\log\frac{p}{q}+(1-p)\log\frac{1-p}{1-q},
 \label{eq:adjustment_bernoulli_kl}
\end{align}
with $0 \cdot \log 0 = 0$.
To assess whether arm $j$ might actually beat $g$, we consider the hypothesis that the true reward mean of $j$ is some $q>\bar r_g$.
The $n_j$ Bernoulli observations of $j$ are best fit by the maximum-likelihood estimate $\bar r_j$, that is, by the empirical distribution $\operatorname{Ber}(\bar r_j)$.
Because $n_j\bar r_j$ of the $n_j$ observations are successes, the log likelihood under mean $q$ is $\ell_j(q)=n_j\{\bar r_j\log q+(1-\bar r_j)\log(1-q)\}$.
Assuming the true mean to be $q$ rather than $\bar r_j$ therefore lowers the log likelihood by
\begin{align}
 \ell_j(\bar r_j)-\ell_j(q)
 &=n_j\Bigl\{\bar r_j\log\frac{\bar r_j}{q}+(1-\bar r_j)\log\frac{1-\bar r_j}{1-q}\Bigr\}\notag\\
 &=n_jd(\bar r_j,q).
\end{align}
The larger this decrease, the more strongly the observations contradict the hypothesis.
Because $\bar r_j<\bar r_g<1$ under our assumptions, $d(\bar r_j,q)$ is continuous and increasing in $q$ on $[\bar r_g,1)$.
By this monotonicity and continuity, the infimum of the KL divergence over all such $q$ is $\inf_{q>\bar r_g}d(\bar r_j,q)=d(\bar r_j,\bar r_g)$.
Line~3 therefore sets $D_j=d(\bar r_j,\bar r_g)=D_{\mathrm{KL}}(\operatorname{Ber}(\bar r_j)\,\|\,\operatorname{Ber}(\bar r_g))$.
This infimum explains why the KL term takes the sample means as its arguments and why $\bar r_j$ comes first.

\subsection{Deriving the Target Relative Exploration Ratio from the IMED Index}
Line~4 converts the KL term into the target relative exploration ratio $\tilde\tau_j$.
For Bernoulli rewards with $\bar r_g<1$, IMED assigns arm $j$ the index $n_jD_j+\log n_j$.
The first term, $n_jD_j$, is the evidence accumulated from the observations of $j$ against alternatives in which $j$ beats $g$.
The second term, $\log n_j$, grows with the number of trials of $j$.
A small index therefore indicates that there is little evidence against $j$ being optimal or that $j$ has few trials.
IMED selects such an arm for further exploration.
We use this principle to design the RS satisfaction level through pairwise comparisons, rather than adopting the IMED selection rule itself.
We define IMED-type indices for $j$ and $g$ as
\begin{align}
 L_j=n_jd(\bar r_j,\bar r_g)+\log n_j,
 \qquad L_g=\log n_g.
 \label{eq:adjustment_imed_index}
\end{align}
(where the KL term for $g$ itself is zero, leaving only $\log n_g$ in its index.)
Following IMED's preference for smaller indices, $L_j\leq L_g$ means that sampling $j$ again has at least as high a priority as sampling $g$ again.
Rewriting this comparison in terms of the actual relative trial proportion $\tau_j=n_j/n_g$ gives
\begin{align}
 L_j\leq L_g
 &\iff n_jd(\bar r_j,\bar r_g)\leq\log\frac{n_g}{n_j}\notag\\
 &\iff \tau_j\leq\exp\{-n_jd(\bar r_j,\bar r_g)\}.
\end{align}
The right-hand side gives a threshold for assessing how extensively arm $j$ has been sampled relative to $g$.
We call this threshold the \textit{target relative exploration ratio} and define it as
\begin{align}
 \tilde\tau_j:=\exp\{-n_jd(\bar r_j,\bar r_g)\}.
 \label{eq:designed_weight}
\end{align}
Because $D_j=d(\bar r_j,\bar r_g)$, Line~4 computes exactly this quantity.
If $\tau_j<\tilde\tau_j$, we treat $j$ as a candidate for further exploration; if $\tau_j>\tilde\tau_j$, we favor exploiting $g$ in this pairwise comparison.
Here, ``target'' refers to the designed switching threshold for the pairwise priority; it does not mean that the actual trial ratio converges to this value.
A unique empirical best arm in an under-achieved condition gives $0\leq\bar r_j<\bar r_g<1$.
Then $d(\bar r_j,\bar r_g)$ is positive because $\bar r_j\neq\bar r_g$, and finite because $0<\bar r_g<1$.
Since $n_j>0$, we obtain $0<\tilde\tau_j<1$.

Constructing the KL term $d(\bar r_j,\bar r_g)$ from the empirical distribution and the largest sample mean provides a plug-in estimate of the corresponding information quantity defined by the true reward distributions and the optimal mean.
The target relative exploration ratio $\tilde\tau_j=\exp\{-n_jd(\bar r_j,\bar r_g)\}$ takes values in $(0,1)$, but it is not a probability.
It is not the posterior probability that arm $j$ is actually optimal.
Nor is it the probability that further trials reverse the order of the sample means so that $\bar r_j$ overtakes $\bar r_g$.
By the Chernoff bound, $\tilde\tau_j$ is only an upper bound on the probability that $n_j$ trials from $\operatorname{Ber}(\bar r_g)$ yield a sample mean of at most $\bar r_j$.
The likelihood relation is an identity for the observed data, whereas recovery from estimation errors depends on the policy.
IMED's performance guarantees apply to its index-minimizing selection rule and do not automatically extend to our RS policy.

\subsection{Translating the Target Ratio into a Pairwise Satisfaction Level}
\label{appsubsec:upper-limit-logic}
Line~5 converts $\tilde\tau_j$ into a pairwise candidate satisfaction level $S_j$.
We translate the switching threshold defined by the target relative exploration ratio into the satisfaction-level scale used by RS.
Specifically, we seek a pairwise satisfaction level that favors further exploration of $j$ when $\tau_j<\tilde\tau_j$ and exploitation of $g$ when $\tau_j>\tilde\tau_j$.
Since RS favors the arm with the larger value, we require the two RS values to tie at the switching threshold $\tau_j=\tilde\tau_j$.
We therefore derive the relation between the satisfaction level and the relative trial proportion from the tie condition $I_g^{\prime}=I_j^{\prime}$.
Substituting the RS-value definition and dividing by $\rho_g>0$ gives (note that $\tau_j = n_j/n_g = \rho_j / \rho_g$)
\begin{align}
 \rho_g(\bar r_g-S^{\prime})&=\rho_j(\bar r_j-S^{\prime}),\notag\\
 \bar r_g-S^{\prime}&=\tau_j(\bar r_j-S^{\prime}).
\end{align}
For $\tau_j\ne1$, solving for the new satisfaction level $S^\prime$ yields
\begin{align}
 S^{\prime}&=\frac{\bar r_g-\tau_j\bar r_j}{1-\tau_j}\notag\\
 &=\bar r_g+\frac{\tau_j}{1-\tau_j}(\bar r_g-\bar r_j).
 \label{eq:satisfaction-level-relational-equation}
\end{align}
For $\bar r_g>\bar r_j$, this tie-inducing satisfaction level lies in the under-achieved range only when $0<\tau_j<1$.
This relation maps each relative trial proportion to the satisfaction level at which the two RS values tie.
We therefore substitute the target relative exploration ratio $\tilde\tau_j\in(0,1)$ for the current ratio $\tau_j$ to define the pairwise candidate satisfaction level:
\begin{align}
 S_j=\bar r_g+
 \frac{\tilde\tau_j}{1-\tilde\tau_j}(\bar r_g-\bar r_j).
 \label{eq:satisfaction-level-relational-equation_with_tau}
\end{align}
Line~5 computes this candidate.
A tie at the threshold alone does not determine the priority on either side, so we next examine the sign of the RS-value difference.
Using $\rho_j=\tau_j\rho_g$, we can write the RS-value difference at $S^{\prime}=S_j$ as
\begin{align}
 I_g^{\prime}-I_j^{\prime}
 =\rho_g\{(\bar r_g-S_j)-\tau_j(\bar r_j-S_j)\}.
\end{align}
From Eq.~\eqref{eq:satisfaction-level-relational-equation_with_tau}, the differences between each sample mean and $S_j$ are
\begin{align}
 \bar r_g-S_j&=-\frac{\tilde\tau_j}{1-\tilde\tau_j}(\bar r_g-\bar r_j),\notag\\
 \bar r_j-S_j&=-\frac{1}{1-\tilde\tau_j}(\bar r_g-\bar r_j).
\end{align}
Substituting these into this difference yields
\begin{align}
 I_g^{\prime}-I_j^{\prime}
 &=\frac{\rho_g(\bar r_g-\bar r_j)}{1-\tilde\tau_j}\{-\tilde\tau_j+\tau_j\}\notag\\
 &=\frac{\rho_g(\bar r_g-\bar r_j)}{1-\tilde\tau_j}
 (\tau_j-\tilde\tau_j).
 \label{eq:adjustment_pairwise_difference}
\end{align}
Because the coefficient is positive, this pairwise comparison favors $j$ simply when $\tau_j<\tilde\tau_j$ and $g$ when $\tau_j>\tilde\tau_j$; the two arms tie only when the ratios are equal.
Thus, setting $S^{\prime}=S_j$ aligns the pairwise RS priority with the comparison between the actual and target trial ratios.

\subsection{Combining Pairwise Levels into a Common Satisfaction Level}
Line~7 combines the pairwise candidates, and Line~8 caps the result at the initial satisfaction level.
To use one satisfaction level for all arms, we take the largest candidate and cap it at the initial satisfaction level:
\begin{align}
 S_{\mathrm{UL}}&:=\max_{j\in J}S_j,\notag\\
 S^{\prime}&:=\min\{S_0 ,S_{\mathrm{UL}}\}.
 \label{eq:actual_satisfaction_level}
\end{align}
Each candidate exceeds $\bar r_g$, so this preserves $\bar r_g<S^{\prime}\leq S_{0}$.
When $\tau_j<\tilde\tau_j$, we have $\rho_g>\rho_j$. In this case, the difference $I_g^{\prime}-I_j^{\prime}$ decreases as the satisfaction level increases.
Consequently, replacing $S_j$ with $S_{\mathrm{UL}}$ preserves the priority of candidate $j$ for further exploration over $g$.
Taking the maximum is conservative in this sense, but it does not guarantee that $j$ will be selected because another arm may have a higher RS value.
This choice also does not preserve every pairwise comparison that favors exploitation or make all pairs tie simultaneously.
The final $S^{\prime}$ preserves the priority of exploration candidates when $S_{0}\geq S_{\mathrm{UL}}$. When $S_{0}<S_{\mathrm{UL}}$, we instead enforce the constraint that the initial satisfaction level must not be increased.
Arms are selected by maximizing RS values at this adjusted satisfaction level, not by minimizing IMED indices; therefore, we make no claim of asymptotic optimality.
Line~9 returns $S^{\prime}$, and RS compares arms at the current step using this adjusted satisfaction level.

\end{document}